%% file: iclr2027_conference.tex
\documentclass{article} 
\usepackage{iclr2027_conference,times}

\input{math_commands.tex}

\usepackage{hyperref}
\usepackage{url}
\usepackage{booktabs}
\usepackage{graphicx}
\usepackage{longtable}

\title{Masking Frequent Tokens Sharpens Direct Preference Optimization}

\author{%
  Harshvardhan Saini\thanks{Equal contribution.} \\
  Indian Institute of Technology Dhanbad\\
  hs1062005@gmail.com\\
  \And
  Samyak Jha\footnotemark[1] \\
  Indian Institute of Technology Dhanbad\\
  samyakjha71@gmail.com\\
  \And
  Yiming Tang\thanks{Corresponding authors.} \\
  National University of Singapore\\
  yiming@nus.edu.sg \\
  \And
  Dianbo Liu\footnotemark[2] \\
  National University of Singapore\\
  dianbo@nus.edu.sg \\
}

\iclrfinalcopy
\begin{document}

\maketitle
\input{sections/abstract}
\input{sections/introduction}
\input{sections/related}

\input{sections/analysis}
\input{sections/method}

\input{sections/results}
\input{sections/conclusion}

\newpage
\subsection*{AI use statement}
In this work, we used generative AI tools, including ChatGPT, Claude, and Claude Code to implement methods, write and edit code, and assist in drafting the manuscript. We did not use generative AI tools to generate synthetic data, formulate , or propose our core analysis, methodology, or experimental design; these were carried out by the authors. We have reviewed all AI-assisted work: code was checked against expected behavior and the underlying data, and all AI-drafted text and result interpretations were verified against experimental outputs and revised where necessary. We take responsibility for the final content of this work, including text, claims, and artifacts produced with the aid of generative AI.

\bibliography{iclr2027_conference}
\bibliographystyle{iclr2027_conference}

\appendix
\input{sections/appendix}

\end{document}

%% file: math_commands.tex
\usepackage{amsmath,amsfonts,bm}

\def\eqref#1{equation~\ref{#1}}

\def\1{\bm{1}}

\DeclareMathAlphabet{\mathsfit}{\encodingdefault}{\sfdefault}{m}{sl}
\SetMathAlphabet{\mathsfit}{bold}{\encodingdefault}{\sfdefault}{bx}{n}



%% file: sections/abstract.tex
\begin{abstract}
Direct Preference Optimization (DPO) aligns language models by optimizing over sequence-level sums of token-wise implicit reward differences. 
However, we identify a pervasive pathology in this formulation: a disproportionately small subset of high-frequency token types dominates cumulative sequence scores while appearing symmetrically across both preferred and dispreferred responses. 
Specifically, under canonical Qwen tokenization on Anthropic HH-RLHF, merely 69 token types account for $55.1\%$ of all response tokens and $85.9\%$ of within-pair shared token mass, exhibiting substantially lower preference-side specificity than the remaining vocabulary. 
This symmetric ubiquity induces gradient entanglement and dilutes the discriminative preference signal propagated through the objective. 
To resolve this issue, we introduce \emph{Anisotropic DPO} (\textsf{ADPO}) and its canonical realization, \emph{Frequency-Hard DPO}. 
Using a fixed, label-agnostic vocabulary mask, our method zeroes the implicit reward contribution of high-frequency response tokens while assigning unit weight to informative positions, thereby suppressing gradient interference without modifying preference pairs, discarding context, or introducing learned parameters. 
Here, \emph{anisotropy} designates non-uniform token-level objective weighting rather than representational geometry. 
Extensive empirical evaluations on AlpacaEval, MT-Bench, and Arena-Hard demonstrate that Frequency-Hard DPO consistently outperforms standard DPO across Qwen-2.5-7B-Instruct and Llama-3-8B-Instruct, establishing that selectively masking shared high-frequency tokens offers an effective, zero-overhead mechanism for robust preference alignment.
\end{abstract}

%% file: sections/introduction.tex
\section{Introduction}
\label{sec:introduction}

Learning from human preferences is a standard approach to aligning language
models \citep{ouyang2022training}. Direct Preference Optimization (DPO)
implements this supervision through a margin between chosen and rejected
responses \citep{rafailov2023direct}. Each response is scored by summing
token-level policy--reference log-ratios with the same explicit coefficient.
Equal coefficients do not imply equal gradients: token contributions depend
on their prefixes and the model's shared parameters. The question is which
of these contributions should directly determine the preference update.

One difficulty is \emph{gradient entanglement}. A margin can improve while
the probabilities of both responses move in the same direction, because
each likelihood change depends on the other response's gradient
\citep{yuan2025common}. Related work documents declining chosen-response
likelihood and displacement toward unintended alternatives
\citep{pal2024smaug,razin2025unintentional}. Shared response content is a
natural place to investigate this coupling. Frequent token types occupy
much of the occurrence mass on both sides of preference pairs, yet show
weaker aggregate chosen--rejected distinction than retained types in our
audits.
This motivates testing whether suppressing their direct score contributions
can reduce the alignment component of gradient entanglement.

\begin{figure*}[t]
\centering
\includegraphics[width=\textwidth]{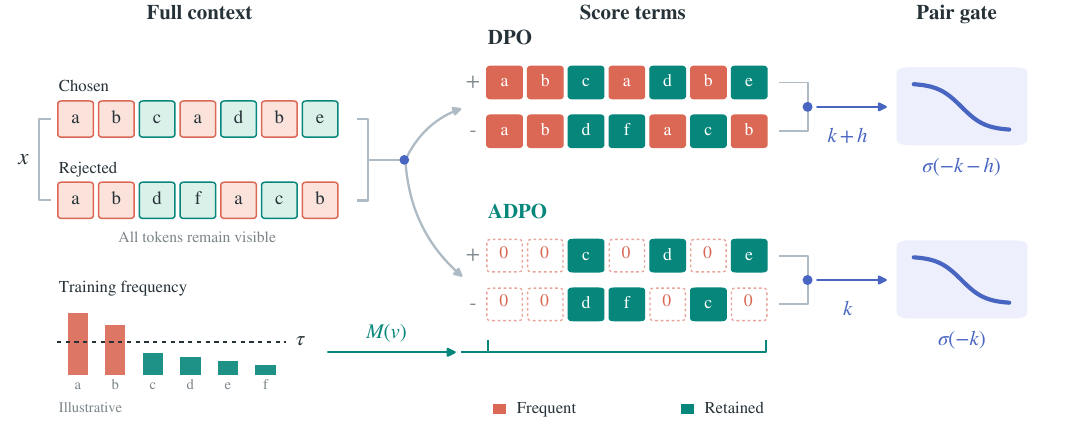}
\caption{\textbf{ADPO masks score terms, not context.}
Coral marks frequent types; teal marks retained types. Letters identify
token types, not score values. Frequencies and coefficients are illustrative,
not a shared positional mask. Each response keeps its full context and
softmax; only coral score terms are zeroed. With beta-scaled retained and
frequent margins $k$ and $h$, selection changes the sigmoid gate from
$\sigma(-k-h)$ to $\sigma(-k)$. Zero direct score weight does not eliminate
indirect gradients through shared parameters.}
\label{fig:adpo_overview}
\end{figure*}

We propose \emph{Anisotropic DPO (ADPO)}, with its canonical form \emph{Frequency-hard
DPO}. ADPO estimates response-token
frequencies once from both sides of training pairs and excludes the
highest-frequency types from the preference score. The mask is fixed,
label-agnostic, and shared by chosen and rejected responses. All tokens
remain in the conditioning context and the full softmax vocabulary.
There is no learned selector, auxiliary reward model, or change to the
architecture. The intervention is \emph{selective regularization} of
preference fitting: it changes both the direct gradient terms and the
pair-level sigmoid multiplier on those that remain
(Figure~\ref{fig:adpo_overview}).

Token reweighting itself is established prior work, including oracle-based,
sparse, and policy-confidence-based approaches
\citep{yang2025selective,christopoulou2024sparsepo,yoon2025confpo}. Our focus is
the simpler corpus-frequency prior. We use \emph{anisotropic preference
optimization} to describe unequal coefficients in the token-score basis, by highlighting that frequent tokens in themselves don't carry any preference signal. 

This work formalizes a novel preference tuning objective by first analyzing chosen-rejected gradient cosine similarities and the preference signals of high-frequency tokens. We isolate genuine predictive information from abundant shared probability mass through structural audits and frozen reward-model comparisons, utilizing random-mask and full-likelihood controls to test the generalizability of our gradient interpretation. Our primary aim is to establish a transparent and computationally inexpensive regularizer for preference fitting. Although we separate the resulting reduction in score-gradient alignment from a general solution to representation entanglement, this targeted restriction consistently yields superior preference optimization.

Our contributions are as follows:
\begin{enumerate}
    \item \textbf{Analysis of frequent-token preference signal.}
    We show that frequent types occupy substantial shared response mass yet exhibit weaker token-identity preference signal in the studied datasets.

    \item \textbf{Analysis of gradient coupling.}
    We characterize how ADPO changes both score-gradient selection and the shared sigmoid gate, and demonstrate reduced chosen--rejected score-gradient alignment at fixed checkpoints.

    \item \textbf{Frequency-based preference optimization.}
    We propose Anisotropic DPO (ADPO), which uses a fixed, training-derived frequency mask to suppress direct preference-score contributions from high-frequency token types while preserving the complete context and full softmax, without a learned selector or auxiliary reward model.
\end{enumerate}

%% file: sections/related.tex
\section{Related Work}
\label{sec:related_work}

\paragraph{Preference learning and direct alignment.}
RLHF learns reward proxies from comparisons and optimizes a policy against
them \citep{stiennon2020learning,ouyang2022training,bai2022training}.
DPO replaces the explicit reward-fitting and RL stages with a
policy--reference preference loss \citep{rafailov2023direct}.
SLiC-HF calibrates sequence likelihoods using feedback
\citep{zhao2023slic}; IPO develops an alternative preference-learning
objective \citep{azar2024general}. ORPO and SimPO remove the separate
reference-model requirement \citep{hong2024orpo,meng2024simpo}, while KTO
uses binary desirability feedback and a prospect-theoretic utility
\citep{ethayarajh2023kto}. SPPO uses a self-play formulation of preference
learning \citep{wu2025self}. ADPO retains DPO's frozen reference and
logistic pairwise loss, changing which realized token terms enter the
margin rather than introducing a new preference model.

\paragraph{Gradient coupling and likelihood displacement.}
DPO-Positive addresses declining chosen-response likelihood, particularly
for pairs with small edit distance \citep{pal2024smaug}.
\citet{yuan2025common} formalize \emph{gradient entanglement}: favorable
likelihood changes depend on the cross-gradient inner product relative to
both squared gradient norms. Their proposed mitigations already include
pairwise gradient normalization and sparsity-regularized contextual token
masking. Separately, \citet{razin2025unintentional} characterize likelihood
displacement through centered hidden-embedding similarity and use it for
data filtering. AlphaPO changes reward shape
\citep{gupta2025alphapo}; normalized-reward objectives regularize the
combined length-normalized probability of the preference pair and examine
outlier-token likelihood changes \citep{im2026normalized}.
ADPO instead tests a fixed frequency selector.

\paragraph{Token-level objectives and credit assignment.}
DPO has a token-level inverse-Q-learning interpretation and can perform
implicit credit assignment \citep{rafailov2023direct}; equal explicit
score coefficients should not be confused with equal token influence. SimPO instead uses the average token log-probability of a response as its
implicit reward, yielding a length-normalized, reference-free
preference objective without an explicit token-level weighting or
reference-model term SimPO \citep{meng2024simpo}. TDPO adds token-level forward-KL control \citep{zeng2024token}. TGDPO introduces token-level reward guidance
\citep{zhu2025tgdpo}. These methods modify the treatment of contextual
token contributions. Selective preference optimization further estimates
token-level reward using a DPO-trained oracle and optimizes only selected
tokens \citep{yang2025selective}, while SparsePO learns sparse token-level
masks to control reward and KL contributions
\citep{christopoulou2024sparsepo}. These methods modify the treatment of contextual
token contributions. ADPO uses no token-reward estimator or importance
sampling correction, and its hard mask deliberately changes the objective
rather than claiming an unbiased estimate of the original DPO gradient.

\paragraph{Frequency priors, nuisance signals, and evaluation.}
Frequency subsampling in word2vec offers a precedent for distinguishing
occurrence count from task usefulness \citep{mikolov2013distributed}, but
does not establish a preference-learning guarantee. Length-aware DPO and
length-controlled evaluation show why apparent preference gains can depend
on response length \citep{park2024disentangling,dubois2024length}.
Reward-model overoptimization further separates proxy fitting from the
desired evaluation target \citep{gao2023scaling}.
Our analyses accordingly distinguish occurrence mass, token-identity
specificity, contextual prediction, and reward-model sensitivity.

\paragraph{Representation anisotropy is a distinct question.}
Representation degeneration \citep{gao2019representation}, anisotropy
from self-attention \citep{godey2024anisotropy}, and output-matrix features
for text embeddings \citep{wu2026your} concern learned
representation geometry. ADPO instead defines anisotropy through unequal
token-score coefficients.

%% file: sections/analysis.tex
\section{Analysis}
\label{sec:main_analysis}

\subsection{The Isotropic Scaling Assumption in DPO}

Consider a prompt $x$ and a response $y=(y_1,\ldots,y_L)$. For each response token $y_t$, define the policy-reference log-ratio as
\begin{equation}
a_t(\theta) = \log \pi_\theta \left( y_t \mid x,y_{<t} \right) - \log \pi_{\mathrm{ref}} \left( y_t \mid x,y_{<t} \right).
\label{eq:token_log_ratio}
\end{equation}
For a chosen response $y^w$ and a rejected response $y^l$, standard DPO constructs the preference margin
\begin{equation}
m(\theta) = \beta \left[ \sum_{t=1}^{L_w} a_t^w(\theta) - \sum_{t=1}^{L_l} a_t^l(\theta) \right],
\label{eq:dpo_margin}
\end{equation}
and minimizes
\begin{equation}
\mathcal{L}_{\mathrm{DPO}} = -\log \sigma\left(m(\theta)\right).
\label{eq:dpo_loss}
\end{equation}
Differentiating the loss with respect to the preference margin gives
\begin{equation}
\frac{\partial \mathcal{L}_{\mathrm{DPO}}}{\partial m} = -\sigma(-m).
\label{eq:dpo_margin_grad}
\end{equation}
Applying the chain rule, the policy gradient is therefore
\begin{equation}
\nabla_\theta \mathcal{L}_{\mathrm{DPO}} = -\sigma(-m)\beta \left[ \sum_{t=1}^{L_w} \nabla_\theta \log \pi_\theta \left( y_t^w \mid x,y_{<t}^w \right) - \sum_{t=1}^{L_l} \nabla_\theta \log \pi_\theta \left( y_t^l \mid x,y_{<t}^l \right) \right].
\label{eq:dpo_parameter_gradient}
\end{equation}
The scalar $\beta$ is therefore applied identically to every realized token. This is the isotropic scaling assumption in DPO: although different tokens may produce very different gradients through the network, their contributions to the preference objective are all multiplied by the same global temperature.

\subsection{A Structural Frequency Prior}
\label{sec:main_frequency_prior}

\begin{figure*}[t]
\centering
\includegraphics[width=\textwidth]{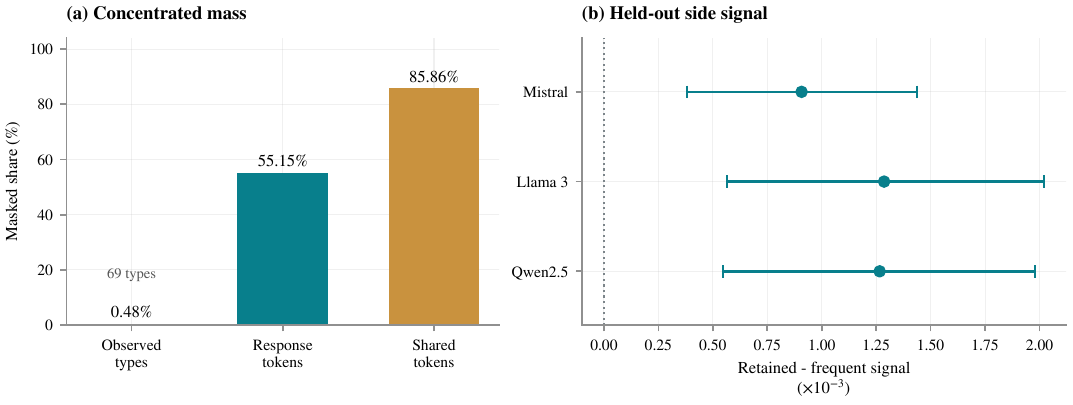}
\caption{\textbf{Frequency-mask coverage and held-out side signal.}
(a) We measure the frequent-type mask's coverage of observed vocabulary types,
response occurrences, and shared chosen--rejected occurrences in HH calibration
data. (b) We compare retained and frequent types using held-out preference-side
associations estimated on separate training pairs. Error bars show paired
confidence intervals. The panels use distinct populations.}
\label{fig:frequency_structure}
\end{figure*}

Frequent types form a small vocabulary subset but account for much of the
overlap between chosen and rejected responses. In most of the textual preference datasets like HH-RLHF
\citep{bai2022training}, their shared occurrence mass is disproportionately
large, while their token-identity association with the preference side is
weaker than that of retained types (Figure~\ref{fig:frequency_structure}(b) and Figure~\ref{fig:rm_preference_signal}(b)).
This motivates a fixed frequency prior for restricting direct preference
fitting without removing the shared context.

We freeze each tokenizer's frequent-type set on training data, then measure
its coverage in disjoint held-out responses from the same
Llama-3-UltraFeedback release \citep{cui2023ultrafeedback,meng2024simpo}.
These types remain common in held-out responses and account for much of their
chosen--rejected overlap.
Separately, Figure~\ref{fig:frequency_structure}(b) measures whether token
associations with the chosen or rejected side repeat on held-out pairs.
Associations estimated on separate training pairs reproduce more strongly
for retained types across the tokenizers examined.

\subsection{Effect of Frequency Masking on Gradient Alignment}
\label{sec:main_gradient_alignment}

Gradient entanglement couples chosen and rejected likelihood changes
\citep{yuan2025common}. We study the effect of frequency masking on chosen--rejected gradient alignment as a diagnostic
for understanding why the frequency-hard objective may behave differently
from standard DPO. We note that we do not intend to claim that frequency masking provides a general mechanism for eliminating gradient entanglement. At each saved DPO checkpoint, we compare full
and frequency-hard score gradients, for $s\in\{w,l\}$:
\begin{equation}
\begin{aligned}
g_s&=\nabla_\theta\sum_t\log\pi_\theta(y_t^s\mid x,y_{<t}^s),\\
a_s&=\nabla_\theta\sum_t M(y_t^s)\log\pi_\theta(y_t^s\mid x,y_{<t}^s).
\end{aligned}
\label{eq:alignment_score_gradients}
\end{equation}
$g_s$ uses all score terms; $a_s$ retains unmasked terms. Both differentiate
all parameters with unchanged context.

\begin{figure*}[t]
\centering
\includegraphics[width=\textwidth]{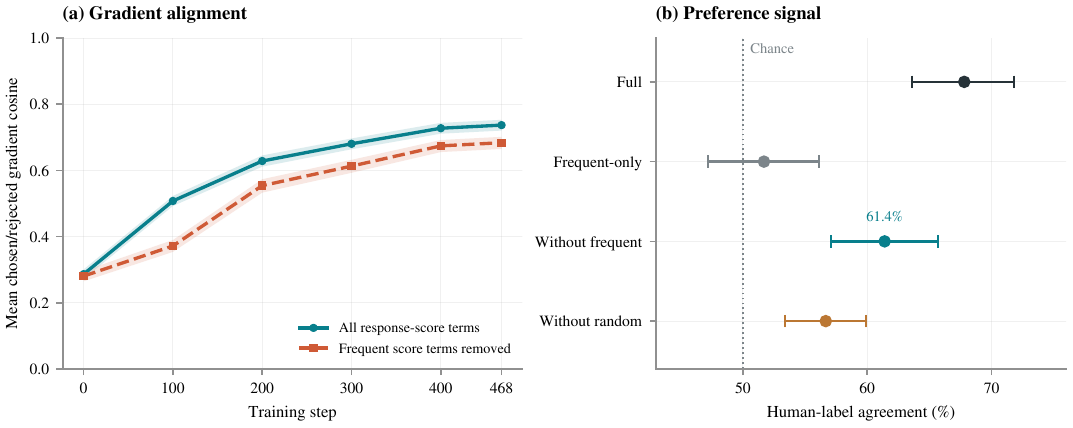}
\caption{\textbf{Gradient alignment and preference signal.}
(a) We compute mean chosen--rejected cosine for full and frequency-hard
response-score gradients at saved DPO checkpoints, using identical weights,
response pairs, and input context.(b) We score complete HH responses, frequent-only
fragments, and frequent- or count-matched random-deletion variants with frozen
PairRM, then compare rankings with the original human labels. Error bars show
prompt-bootstrap intervals, and the vertical line marks chance. The panels
use different samples and interventions: score masking versus response editing.}
\label{fig:gradient_frequency_masking_dpo}
\label{fig:rm_preference_signal}
\end{figure*}

Figure~\ref{fig:gradient_frequency_masking_dpo}(a) uses identical weights and
response pairs within each comparison. Removing frequent score terms lowers
mean cosine at every measured checkpoint, suggesting reduced chosen--rejected
directional coupling under frequency-hard scoring throughout the sampled
trajectory. This observation is consistent with frequency
masking changing which token-level score contributions participate in the
gradient, providing a diagnostic for a possible mechanism underlying the
empirical behavior of the frequency-hard objective.

\subsection{Preference Signal in Frequent Tokens}
\label{sec:main_reward_sensitivity}
\label{sec:main_reward_fragments}

The frequency-hard mask is motivated by the hypothesis that highly frequent
tokens carry less preference-relevant information than the tokens it retains.
We therefore evaluate the preference signal associated with frequent-token
content using both a frozen reward model and human preference labels.

On a held-out HH sample, we construct four response conditions: the complete
response, a frequent-only fragment containing the masked token types, the
complement obtained by removing frequent tokens, and a count-matched random
deletion control. We score all conditions using frozen PairRM
\citep{jiang2023llm} and compare the resulting rankings against the original
human preference labels. As shown in Figure~\ref{fig:rm_preference_signal}(b)
and Table~\ref{tab:rm_frequency_fragments}, the frequent-only fragments
produce a near-zero reward margin ($0.08779$) and near-chance agreement with
human labels ($51.7\%$). In contrast, removing the frequent tokens retains a
substantially larger reward margin ($0.97335$) and higher human-label
agreement ($61.4\%$).

The gap between complete and edited responses can
naturally reflect the loss of response length, context, and coherence
introduced by the editing procedure, which limits what can be inferred from
the absolute performance of the fragment conditions. Our central observation
is instead that the non-frequent complement retains substantially more
preference signal than the frequent-token fragment. The count-matched random
deletion condition provides a supplementary control for the effect of
removing a comparable number of tokens.

Taken together, these results suggest that measurable preference signal is
concentrated more strongly in the non-frequent content than in the
frequent-token fragments, providing empirical motivation for restricting
direct DPO fitting on the frequent token types while retaining them in the
causal context.

\begin{table*}[t]
\centering
\small
\setlength{\tabcolsep}{4pt}
\caption{\textbf{Human-label agreement and reward margins.}
We evaluate the response conditions in Figure~\ref{fig:rm_preference_signal}(b)
against the original HH preference labels. Ties receive half credit;
margins are order-symmetrized raw PairRM logits, not reward probabilities.}
\label{tab:rm_frequency_fragments}
\begin{tabular}{lrr}
\toprule
Condition & Agreement (\%) & Margin \\
\midrule
Full & 67.8 & 1.89969 \\
Frequent-only & 51.7 & 0.08779 \\
Without frequent & 61.4 & 0.97335 \\
Without random & 56.67 & 0.60148 \\
\bottomrule
\end{tabular}
\end{table*}

%% file: sections/method.tex
\section{Anisotropic DPO}
\label{sec:methodology}

\emph{Anisotropic DPO (ADPO)} which is also the \emph{Frequency-hard DPO} name the same
single method: fixed binary weighting of realized token-score terms using
empirical response frequencies.

\subsection{The DPO Token Score}
\label{sec:adpo_preliminaries}

For preference pair $i$, let $y_i^w$ and $y_i^l$ denote the chosen and
rejected responses. With the reference policy fixed, define
\begin{equation}
\begin{aligned}
a_{it}^s(\theta)
={}&\log\pi_\theta(y_{it}^s\mid x_i,y_{i,<t}^s)\\
&-\log\pi_{\mathrm{ref}}(y_{it}^s\mid x_i,y_{i,<t}^s),
\quad s\in\{w,l\}.
\end{aligned}
\label{eq:token_log_ratio_freqhard}
\end{equation}
The ordinary DPO margin and objective are
\begin{equation}
m_i(\theta)=\beta\left[
\sum_{t=1}^{L_i^w}a_{it}^w(\theta)
-\sum_{t=1}^{L_i^l}a_{it}^l(\theta)
\right],
\label{eq:dpo_margin_freqhard}
\end{equation}
\begin{equation}
\mathcal L_{\mathrm{DPO}}(\theta)
=-\frac{1}{N}\sum_{i=1}^N\log\sigma(m_i(\theta)).
\label{eq:dpo_loss_freqhard}
\end{equation}
Our intervention changes the terms included in this score, while retaining
the logistic pairwise loss and frozen reference.

\subsection{The Fixed Frequency Mask}
\label{sec:frequency_estimation}
\label{sec:frequency_hard_mask}

For each tokenizer, count response positions in a training-calibration subset
$\mathcal D_f$ of $M$ pairs, using the training tokenization, truncation, and
EOS handling. The count and normalized frequency of type $v$ are
\begin{equation}
c(v)=\sum_{i=1}^{M}\left[c_i^w(v)+c_i^l(v)\right],
\qquad
\widehat f(v)=\frac{c(v)}{\sum_{u\in\mathcal V}c(u)}.
\label{eq:freqhard_frequency}
\end{equation}
Both sides enter symmetrically, making selection invariant to swapping the
preference labels. The canonical configuration uses the first $M=4{,}000$
training pairs, the model's chat template, prompt/sequence caps of 256/512
tokens, and response EOS. Counts are recomputed for each tokenizer.

For threshold $\tau$, define the vocabulary-level weight
\begin{equation}
w_\tau(v)=\mathbf{1}[\widehat f(v)\leq\tau].
\label{eq:frequency_hard_weight}
\end{equation}
We use the strict threshold $\tau=0.002$, retain unseen types at unit weight,
and fix $w_\tau$ throughout training. This is the mask $M$ in the analysis.
Write $\mathcal H_\tau=\{v:\widehat f(v)>\tau\}$ and
$\mathcal K_\tau=\mathcal V\setminus\mathcal H_\tau$ for the masked and
retained vocabulary sets. We have provided an ablation on $\tau$ later.

\subsection{The Frequency-Hard Objective}
\label{sec:frequency_hard_objective}

Using Equation~\ref{eq:token_log_ratio_freqhard}, the selected sequence score is
\begin{equation}
R_{\theta,\tau}(x_i,y_i^s)
=\sum_{t=1}^{L_i^s}w_\tau(y_{it}^s)a_{it}^s(\theta).
\label{eq:frequency_hard_sequence_score}
\end{equation}
Frequency-hard DPO minimizes
\begin{equation}
\begin{aligned}
 m_{i,\tau}^{\mathrm{FH}}(\theta)
 &=\beta\left[
R_{\theta,\tau}(x_i,y_i^w)-R_{\theta,\tau}(x_i,y_i^l)
\right],\\
\mathcal L_{\mathrm{FH}}(\theta)
&=-\frac{1}{N}\sum_{i=1}^{N}
\log\sigma\!\left(m_{i,\tau}^{\mathrm{FH}}(\theta)\right).
\end{aligned}
\label{eq:frequency_hard_loss}
\end{equation}
The effective coefficient is $\beta_{it}=\beta\,w_\tau(y_{it})$.
With no masked realized types, the objective reduces exactly to standard DPO.

Scores are not renormalized by retained token count. A response with no
retained positions has zero direct score; if both sides have none, the pair
has loss $\log 2$ and zero direct gradient, and is not dropped. All original
tokens remain in the causal context and log-probabilities use the full softmax
vocabulary. Zero weight is not a frozen output row or an attention mask:
shared parameters can still receive indirect gradients.

Writing the full beta-scaled margin as retained $k$ plus frequent $h$, ADPO
changes the negative loss gradient from $\sigma(-k-h)\nabla(k+h)$ to
$\sigma(-k)\nabla k$. It therefore combines score-gradient selection with
pair-dependent gate reweighting.

\begin{table}[htbp]
\centering
\small
\renewcommand{\arraystretch}{1.15}
\caption{
Token types removed by the frequency-hard mask at $\tau=0.002$.
The mask contains 50 vocabulary types whose empirical frequency exceeds
the threshold.
}
\label{tab:frequency_hard_tokens}

\begin{tabular}{p{0.29\linewidth} p{0.64\linewidth}}
\toprule
\textbf{Token type} & \textbf{Tokens masked} \\
\midrule

Function words / stopwords
& \texttt{the}, \texttt{is}, \texttt{The}, \texttt{and}, \texttt{for},
\texttt{to}, \texttt{that}, \texttt{a},
\texttt{with}, \texttt{it}, \texttt{of}, \texttt{or}, \texttt{are},
\texttt{in}, \texttt{can}, \texttt{be}, \texttt{on}, \texttt{by},
\texttt{as} \\

Punctuation
& \texttt{,}, \texttt{.}, \texttt{:}, \texttt{)}, \texttt{"} \\

Markdown / code markup
& \texttt{*}, \texttt{**}, \texttt{**:}, \texttt{**},
\texttt{\textasciigrave}, \texttt{\textasciigrave\textasciigrave} \\

Whitespace / Newline composites
& \textit{space}, \texttt{\textbackslash n},
\textit{three-space indent}, \texttt{\textbackslash t},
\texttt{.}\texttt{\textbackslash n},
\texttt{.}\texttt{\textbackslash n\textbackslash n},
\texttt{:}\texttt{\textbackslash n\textbackslash n},
\texttt{**}\texttt{\textbackslash n\textbackslash n} \\

\bottomrule
\end{tabular}
\end{table}

%% file: sections/results.tex
\section{Results}
\label{sec:results}

\newcommand{\result}[2]{#1\,{\scriptsize$\pm$#2}}

\begin{table}[htbp]
\centering
\small
\renewcommand{\arraystretch}{1.15}
\caption{
Main benchmark results for Qwen2.5-7B-Instruct.
MT-Bench and AlpacaEval 2.0 use GPT-4-turbo as the API judge, while
Arena-Hard uses GPT-4.1-mini as the API judge.
$\pm$ values denote respective uncertainity estimates.
}
\vspace{4pt}
\label{tab:qwen_main_results}

\begin{tabular}{lcccc}
\toprule
\parbox[c][2.8\baselineskip][c]{1.2cm}{\centering\textbf{Method}}
& \textbf{FsFairX RM}
& \textbf{MT-Bench}
& \textbf{AlpacaEval 2.0}
& \textbf{Arena-Hard} \\
& \textbf{Mean Reward}
& \textbf{Score}
& \textbf{LC Win Rate}
& \textbf{LC Win Rate} \\
\midrule

DPO
& \result{1.231}{0.045}
& \result{8.43}{0.060}
& \result{49.72\%}{0.65\%}
& \result{52.1\%}{0.65\%} \\

SimPO
& \result{1.180}{0.055}
& \result{8.15}{0.072}
& \result{47.80\%}{0.72\%}
& \result{51.6\%}{0.69\%} \\

TGDPO
& \result{1.275}{0.045}
& \result{8.42}{0.060}
& \result{51.40\%}{0.65\%}
& \result{52.8\%}{0.64\%} \\

TDPO
& \result{1.220}{0.051}
& \result{8.475}{0.055}
& \result{49.90\%}{0.71\%}
& \result{53.0\%}{0.55\%} \\

SparsePO
& \result{1.255}{0.053}
& \result{8.46}{0.061}
& \result{51.10\%}{0.63\%}
& \result{52.6\%}{0.51\%} \\

SePO
& \result{1.245}{0.055}
& \result{8.40}{0.065}
& \result{50.60\%}{0.68\%}
& \result{52.3\%}{0.65\%} \\

\textbf{ADPO}
& \textbf{\result{1.308}{0.052}}
& \textbf{\result{8.53}{0.057}}
& \textbf{\result{52.78\%}{0.46\%}}
& \textbf{\result{53.4\%}{0.59\%}} \\

\bottomrule
\end{tabular}
\end{table}

\begin{table}[htbp]
\centering
\small
\renewcommand{\arraystretch}{1.15}
\caption{
Main benchmark results for Llama-3-8B-Instruct.
MT-Bench and AlpacaEval 2.0 use GPT-4-turbo as the API judge, while
Arena-Hard uses GPT-4.1-mini as the API judge.
}
\vspace{4pt}
\label{tab:llama_main_results}

\begin{tabular}{lcccc}
\toprule
\parbox[c][2.8\baselineskip][c]{1.2cm}{\centering\textbf{Method}}
& \textbf{FsFairX RM}
& \textbf{MT-Bench}
& \textbf{AlpacaEval 2.0}
& \textbf{Arena-Hard} \\
& \textbf{Mean Reward}
& \textbf{Score}
& \textbf{LC Win Rate}
& \textbf{LC Win Rate} \\
\midrule

DPO
& \result{2.15}{0.058}
& \result{7.66}{0.067}
& \result{50.43\%}{0.72\%}
& \result{34.9\%}{0.68\%} \\

SimPO
& \result{1.95}{0.063}
& \result{7.60}{0.074}
& \result{48.50\%}{0.77\%}
& \result{32.8\%}{0.73\%} \\

TGDPO
& \result{2.17}{0.052}
& \result{7.75}{0.069}
& \result{52.00\%}{0.64\%}
& \result{35.2\%}{0.62\%} \\

TDPO
& \result{2.14}{0.057}
& \result{7.85}{0.061}
& \result{50.00\%}{0.71\%}
& \result{35.0\%}{0.66\%} \\

SparsePO
& \result{2.17}{0.054}
& \result{7.80}{0.068}
& \result{51.40\%}{0.67\%}
& \result{35.4\%}{0.59\%} \\

SePO
& \result{2.13}{0.061}
& \result{7.72}{0.063}
& \result{50.70\%}{0.69\%}
& \result{35.1\%}{0.72\%} \\

\textbf{ADPO}
& \textbf{\result{2.20}{0.059}}
& \textbf{\result{7.89}{0.066}}
& \textbf{\result{53.06\%}{0.63\%}}
& \textbf{\result{36.5\%}{0.48\%}} \\

\bottomrule
\end{tabular}
\end{table}

\subsection{Experimental Setup}
\label{sec:experimental_setup}
We evaluate the proposed framework on Llama-3-8B-Instruct, and Qwen2.5-7B-Instruct. For each model, the policy is initialized from the corresponding instruction-tuned checkpoint, while an identical frozen copy is used as the reference model (wherever applicable). No supervised fine-tuning stage is performed before preference optimization.

To construct the preference data, we follow the on-policy data generation
recipe introduced by SimPO. Starting from the
UltraFeedback prompts, we regenerate candidate responses using the
corresponding instruction-tuned model rather than using the original
preference responses. For each prompt, we sample five responses from the
model with a temperature of $0.8$ and use RLHFlow/ArmoRM-Llama3-8B-v0.1 \citep{ArmoRM} to
rank them. The highest-scoring response is selected as the chosen response
$y^w$, while the lowest-scoring response is selected as the rejected
response $y^l$. We generate the preference data in a single pass, following
the SimPO procedure, thereby making the preference distribution closer to
that of the policy being optimized.

For the main evaluations, methods are trained via Full Sharded Data Parallel (FSDP) fine-tuning. Across all evaluated models, the global batch size is configured to 128 (using a per-device batch size of 4, gradient accumulation of 16, and 2 GPUs). The training runs for 1 epoch using bf16 precision, with a maximum sequence length of 2048. 

We perform a hyperparameter sweep over the DPO temperature $\beta$ for all
methods except SimPO, evaluating $\beta\in\{0.01,0.05,0.1\}$. For SimPO, which
uses a different reward scaling, we separately evaluate larger $\beta$ values
and select the best-performing setting. The hyperparameters reported below
correspond to the selected configurations used for the main benchmark
results in Tables~\ref{tab:qwen_main_results} and
\ref{tab:llama_main_results}.
\begin{itemize}
    \item \textbf{DPO}: Serves as the reference implementation, utilizing a temperature parameter $\beta=0.01$ and learning rate $5\times10^{-7}$.
    \item \textbf{SimPO}: Uses a reference-free length-normalized reward with $\beta=2.5$, a target margin $\gamma=1.375$ (calculated as $0.55\cdot\beta$), and learning rate $1\times10^{-6}$.
    \item \textbf{TGDPO}: Employs a per-token forward-KL regularizer using $\beta=0.01$, a token reward factor $\alpha=0.5$, and learning rate $5\times10^{-7}$.
    \item \textbf{TDPO}: Utilizes $\beta=0.1$, a KL weight $\alpha=0.5$, and learning rate $5\times10^{-7}$.
    \item \textbf{SparsePO}: Uses a token-level reward and KL mask with $\beta=0.1$, learning rate $5\times10^{-7}$, mask weight decay $0.01$, and L1 regularization $0.001$ on the mask weights.
    \item \textbf{SePO}: Uses the top-$30\%$ token-selection setting and a reference-free contrastive objective; the policy is trained with learning rate $5\times10^{-7}$.
    \item \textbf{ADPO}: Our proposed frequency-hard token-weighted DPO, applied with $\beta=0.01$, learning rate $5\times10^{-7}$, a frequency mask threshold $\tau=0.002$.
\end{itemize}

\begin{table}[htbp]
\centering
\small
\renewcommand{\arraystretch}{1.15}
\caption{
Sensitivity of ADPO to the frequency threshold $\tau$ on
Llama-3-8B-Instruct. Each threshold is selected to mask approximately
the indicated number of vocabulary tokens. FsFairX RM reports the mean
reward assigned by the FsFairX Llama-3 RM v0.1, while AlpacaEval 2.0
reports length-controlled (LC) win rate.
}
\label{tab:tau_ablation}

\begin{tabular}{cccc}
\toprule
\textbf{$\tau$}
& \textbf{Masked Tokens}
& \textbf{FsFairX RM}
& \textbf{AlpacaEval 2.0} \\
& \textbf{(Approx.)}
& \textbf{Mean Reward}
& \textbf{LC Win Rate} \\
\midrule

0.00413
& 25
& 1.654
& 48.15\% \\

0.00200
& 50
& \textbf{2.200}
& \textbf{53.06\%} \\

0.00084
& 100
& 1.440
& 42.70\% \\

\bottomrule
\end{tabular}
\end{table}

\begin{table}[htbp]
\centering
\small
\renewcommand{\arraystretch}{1.15}
\caption{
Ablation of token-selection strategies on Llama-3-8B-Instruct using
approximately 50 token types. Hard-mask variants remove the indicated
token set from the preference score while preserving the full context.
Soft inverse-frequency weighting assigns larger weights to less frequent
token types.
}
\label{tab:token_selection_ablation}

\begin{tabular}{lcc}
\toprule
\textbf{Method}
& \textbf{FsFairX RM}
& \textbf{AlpacaEval 2.0} \\
& \textbf{Mean Reward}
& \textbf{LC Win Rate} \\
\midrule

DPO
& 2.15
& 50.43\% \\

Random mask (50 tokens)
& 2.14
& 50.18\% \\

Mid-frequency mask (50 tokens)
& 2.11
& 49.72\% \\

Low-frequency mask (50 tokens)
& 2.14
& 49.55\% \\

Soft inverse-frequency weights
& 2.16
& 50.62\% \\

\textbf{ADPO}
& \textbf{2.20}
& \textbf{53.06\%} \\

\bottomrule
\end{tabular}
\end{table}

\subsection{Main Results}
\label{sec:main_results}

Tables~3 and~4 compare Frequency-hard DPO (ADPO) with the considered
baselines on Qwen2.5-7B-Instruct and Llama-3-8B-Instruct. On
Qwen2.5-7B-Instruct, ADPO increases the FsFairX mean reward from 1.231 to
1.308 and the MT-Bench score from 8.43 to 8.53 relative to DPO.
AlpacaEval~2.0 length-controlled win rate improves from 49.72\% to
52.78\%, while the reported Arena-Hard win rate increases from 52.1\% to
53.4\%. ADPO achieves the highest reported score on each metric among the
compared methods.

On Llama-3-8B-Instruct, ADPO improves the FsFairX mean reward from 2.15 to
2.20 and the MT-Bench score from 7.66 to 7.89. AlpacaEval~2.0
length-controlled win rate increases from 50.43\% to 53.06\%, and the
reported Arena-Hard win rate increases from 34.9\% to 36.5\%. ADPO achieves
the highest reported score on each metric among the compared methods.

Table~5 examines sensitivity to the frequency threshold $\tau$ on
Llama-3-8B-Instruct. Among the tested settings, $\tau=0.002$, masking
approximately 50 token types, achieves the highest mean reward and
AlpacaEval~2.0 length-controlled win rate. Masking approximately 25 or 100
tokens yields lower scores on both metrics. The token-selection ablation shows that none of the alternative selection strategies provides a benefit over the standard DPO objective.

%% file: sections/conclusion.tex
\section{Conclusion}
\label{sec:conclusion}
This work demonstrates that the isotropic scaling assumption in standard Direct Preference Optimization inadvertently allows a small subset of high-frequency tokens to dominate and entangle preference gradients. By introducing Anisotropic DPO (ADPO), we establish that selectively masking these frequent token types from the direct preference score---without discarding context, modifying pairs, or introducing learned parameters---effectively mitigates this interference. Extensive evaluations on Qwen2.5-7B-Instruct and Llama-3-8B-Instruct confirm that this zero-overhead frequency prior consistently improves alignment outcomes across multiple rigorous benchmarks, offering a highly efficient and transparent mechanism for robust language model preference optimization.

%% file: sections/appendix.tex
\section{Appendix}

\subsection{Frequency Structure Across Tokenizers}
\label{app:adpo_frequency_structure}

\paragraph{Audit design.}
We evaluate the frequency prior across Mistral, Llama~3, and Qwen2.5
tokenizers using the same released Llama-3-UltraFeedback response pairs.
For each tokenizer, the first 4,000 training pairs determine the fixed
frequent-type set
\begin{equation}
\mathcal H_\tau=\{v:\widehat f(v)>\tau\},\qquad
\mathcal K_\tau=\mathcal V\setminus\mathcal H_\tau,\qquad
\tau=0.002.
\label{eq:app_adpo_frequency_partition}
\end{equation}
Both responses contribute symmetrically to $\widehat f$.
The audit applies each tokenizer's chat template, includes response EOS,
and uses sequence and prompt caps of 2,048 and 1,800 tokens, respectively.
These are the settings of this structural audit. The selected token IDs
are frozen before evaluation on all 1,961 disjoint test pairs.
Separate training pairs 4,001--8,000 calibrate the signed side associations
used below. This design separates mask construction, association
calibration, and held-out evaluation.

\paragraph{Coverage and shared occurrence mass.}
Let $c_i^s(v)$ count type $v$ in response side $s\in\{w,l\}$, and let
$n_{\mathcal D}(v)=\sum_{i\in\mathcal D}[c_i^w(v)+c_i^l(v)]$.
For a type set $\mathcal S$, define its occurrence coverage and shared
occurrence coverage on population $\mathcal D$ as
\begin{align}
F_{\mathcal D}(\mathcal S)
&=\frac{\sum_{v\in\mathcal S}n_{\mathcal D}(v)}
{\sum_{v\in\mathcal V}n_{\mathcal D}(v)},\\
J_{\mathcal D}(\mathcal S)
&=\frac{\sum_{i\in\mathcal D}\sum_{v\in\mathcal S}
\min\{c_i^w(v),c_i^l(v)\}}
{\sum_{i\in\mathcal D}\sum_{v\in\mathcal V}
\min\{c_i^w(v),c_i^l(v)\}}.
\label{eq:app_adpo_shared_coverage}
\end{align}
For side-normalized rates
$p_{s,\mathcal D}(v)=\sum_{i\in\mathcal D}c_i^s(v)/
\sum_{i\in\mathcal D}\sum_u c_i^s(u)$, define
\begin{equation}
q_{\mathcal D}(v)=
\frac{|p_{w,\mathcal D}(v)-p_{l,\mathcal D}(v)|}
{p_{w,\mathcal D}(v)+p_{l,\mathcal D}(v)},\qquad
\overline q_{\mathcal D}(\mathcal S)=
\frac{\sum_{v\in\mathcal S}n_{\mathcal D}(v)q_{\mathcal D}(v)}
{\sum_{v\in\mathcal S}n_{\mathcal D}(v)}.
\label{eq:app_adpo_side_specificity}
\end{equation}
Unobserved types contribute zero. The occurrence-weighted statistic
$\overline q$ measures absolute token-identity side specificity.

Table~\ref{tab:app_adpo_tokenizer_structure} establishes the same
structural pattern across all three tokenizers: a small frequent-type set
captures a disproportionate share of chosen--rejected overlap and has
substantially weaker token-identity side specificity than its complement.
The masks cover 36.06--40.94\% of held-out response occurrences while
capturing 47.43--52.49\% of their shared occurrence mass.

\paragraph{Reproducible preference-side associations.}
To measure whether side associations repeat out of sample, define the
signed calibration skew
\begin{equation}
d_{\mathrm{cal}}(v)=
\frac{p_{w,\mathrm{cal}}(v)-p_{l,\mathrm{cal}}(v)}
{p_{w,\mathrm{cal}}(v)+p_{l,\mathrm{cal}}(v)},
\end{equation}
setting it to zero for types absent from the association-calibration
population. The held-out cross-fitted score is
\begin{equation}
C(\mathcal S)=
\frac{\sum_{v\in\mathcal S}d_{\mathrm{cal}}(v)
[p_{w,\mathrm{test}}(v)-p_{l,\mathrm{test}}(v)]}
{\sum_{v\in\mathcal S}
[p_{w,\mathrm{test}}(v)+p_{l,\mathrm{test}}(v)]}.
\label{eq:app_adpo_crossfit_signal}
\end{equation}
Positive values indicate that the direction of a token's side association
reproduces on disjoint pairs. We resample evaluation preference pairs
jointly across the frequent and retained groups for 10,000 bootstrap
draws, keeping the training-derived mask and calibration associations
fixed. The resulting intervals quantify evaluation-pair uncertainty.

\begin{table}[t]
\centering
\small
\setlength{\tabcolsep}{5pt}
\caption{Held-out coverage and side specificity of training-derived
frequency masks. Coverage is computed from response occurrences; shared
coverage uses within-pair multiset overlap. The same test pairs are
tokenized separately for each row.}
\label{tab:app_adpo_tokenizer_structure}
\begin{tabular}{lrrrrr}
\toprule
Tokenizer & $|\mathcal H_\tau|$ & $F(\mathcal H_\tau)$ (\%) &
$J(\mathcal H_\tau)$ (\%) & $\overline q(\mathcal H_\tau)$ &
$\overline q(\mathcal K_\tau)$ \\
\midrule
Mistral & 53 & 40.94 & 52.49 & 0.0126 & 0.1033 \\
Llama~3 & 50 & 36.06 & 47.43 & 0.0133 & 0.1406 \\
Qwen2.5 & 52 & 37.77 & 49.22 & 0.0135 & 0.1404 \\
\bottomrule
\end{tabular}
\end{table}

\begin{table}[t]
\centering
\small
\setlength{\tabcolsep}{7pt}
\caption{Cross-fitted preference-side signal. All entries are scaled by
$10^3$. Intervals are paired percentile bootstrap intervals for the
retained-minus-frequent difference, computed separately per tokenizer.}
\label{tab:app_adpo_crossfit_signal}
\begin{tabular}{lrrrr}
\toprule
Tokenizer & $C(\mathcal H_\tau)$ & $C(\mathcal K_\tau)$ & Difference & 95\% CI \\
\midrule
Mistral & 0.326 & 1.234 & 0.908 & $[0.382,1.439]$ \\
Llama~3 & 0.315 & 1.602 & 1.287 & $[0.563,2.022]$ \\
Qwen2.5 & 0.292 & 1.558 & 1.266 & $[0.546,1.981]$ \\
\bottomrule
\end{tabular}
\end{table}

Retained types have a larger cross-fitted side signal under every
tokenizer, and each paired interval for the difference is above zero
(Table~\ref{tab:app_adpo_crossfit_signal}). Together, the coverage and
cross-fitted analyses support a tokenizer-replicated frequency prior:
frequent types concentrate shared occurrences, whereas retained types
carry stronger reproducible token-identity preference-side associations
per unit group mass.

\paragraph{Exact masked-token inventories.}
Table~\ref{tab:app_adpo_token_inventory} gives every selected token ID and
its tokenizer-piece string for the three training-derived masks. Strings
are JSON-escaped, with non-ASCII characters represented by Unicode escapes.
For example, \texttt{\textbackslash{}u2581} is the Mistral word-boundary
marker; \texttt{\textbackslash{}u0120},
\texttt{\textbackslash{}u010a}, and \texttt{\textbackslash{}u0109} denote
the byte-level space, newline, and tab markers in the other tokenizers.
These are vocabulary pieces, not separately decoded text fragments.
Each tokenizer's columns follow descending calibration frequency;
positions across tokenizers do not imply corresponding token identities.

\begingroup
\scriptsize
\setlength{\tabcolsep}{3.5pt}
\begin{longtable}{rlrlrl}
\caption{Complete frequent-type inventories for the structural audit.
Token IDs and escaped vocabulary pieces preserve whitespace, byte, and
special-token distinctions.}
\label{tab:app_adpo_token_inventory}\\
\toprule
\multicolumn{2}{c}{Mistral} & \multicolumn{2}{c}{Llama~3} &
\multicolumn{2}{c}{Qwen2.5} \\
ID & Vocabulary piece & ID & Vocabulary piece & ID & Vocabulary piece \\
\midrule
\endfirsthead
\multicolumn{6}{c}{Table~\thetable\ continued}\\
\toprule
\multicolumn{2}{c}{Mistral} & \multicolumn{2}{c}{Llama~3} &
\multicolumn{2}{c}{Qwen2.5} \\
ID & Vocabulary piece & ID & Vocabulary piece & ID & Vocabulary piece \\
\midrule
\endhead
\midrule
\multicolumn{6}{r}{Continued on the next page}\\
\endfoot
\bottomrule
\endlastfoot
13 & \texttt{"<0x0A>"} & 11 & \texttt{","} & 11 & \texttt{","} \\
28723 & \texttt{"."} & 279 & \texttt{"\textbackslash{}u0120the"} & 279 & \texttt{"\textbackslash{}u0120the"} \\
28725 & \texttt{","} & 323 & \texttt{"\textbackslash{}u0120and"} & 323 & \texttt{"\textbackslash{}u0120and"} \\
272 & \texttt{"\textbackslash{}u2581the"} & 13 & \texttt{"."} & 13 & \texttt{"."} \\
304 & \texttt{"\textbackslash{}u2581and"} & 311 & \texttt{"\textbackslash{}u0120to"} & 311 & \texttt{"\textbackslash{}u0120to"} \\
298 & \texttt{"\textbackslash{}u2581to"} & 264 & \texttt{"\textbackslash{}u0120a"} & 264 & \texttt{"\textbackslash{}u0120a"} \\
264 & \texttt{"\textbackslash{}u2581a"} & 315 & \texttt{"\textbackslash{}u0120of"} & 315 & \texttt{"\textbackslash{}u0120of"} \\
28705 & \texttt{"\textbackslash{}u2581"} & 220 & \texttt{"\textbackslash{}u0120"} & 220 & \texttt{"\textbackslash{}u0120"} \\
302 & \texttt{"\textbackslash{}u2581of"} & 25 & \texttt{":"} & 16 & \texttt{"1"} \\
28747 & \texttt{":"} & 304 & \texttt{"\textbackslash{}u0120in"} & 15 & \texttt{"0"} \\
28740 & \texttt{"1"} & 9 & \texttt{"*"} & 25 & \texttt{":"} \\
28742 & \texttt{"'"} & 627 & \texttt{".\textbackslash{}u010a"} & 17 & \texttt{"2"} \\
28733 & \texttt{"-"} & 374 & \texttt{"\textbackslash{}u0120is"} & 304 & \texttt{"\textbackslash{}u0120in"} \\
28734 & \texttt{"0"} & 198 & \texttt{"\textbackslash{}u010a"} & 9 & \texttt{"*"} \\
28713 & \texttt{"s"} & 3146 & \texttt{"\textbackslash{}u0120**"} & 624 & \texttt{".\textbackslash{}u010a"} \\
297 & \texttt{"\textbackslash{}u2581in"} & 320 & \texttt{"\textbackslash{}u0120("} & 374 & \texttt{"\textbackslash{}u0120is"} \\
28750 & \texttt{"2"} & 382 & \texttt{".\textbackslash{}u010a\textbackslash{}u010a"} & 198 & \texttt{"\textbackslash{}u010a"} \\
28736 & \texttt{"*"} & 430 & \texttt{"\textbackslash{}u0120that"} & 3070 & \texttt{"\textbackslash{}u0120**"} \\
349 & \texttt{"\textbackslash{}u2581is"} & 369 & \texttt{"\textbackslash{}u0120for"} & 320 & \texttt{"\textbackslash{}u0120("} \\
325 & \texttt{"\textbackslash{}u2581("} & 596 & \texttt{"'s"} & 382 & \texttt{".\textbackslash{}u010a\textbackslash{}u010a"} \\
619 & \texttt{"\textbackslash{}u2581**"} & 449 & \texttt{"\textbackslash{}u0120with"} & 429 & \texttt{"\textbackslash{}u0120that"} \\
348 & \texttt{"**"} & 330 & \texttt{"\textbackslash{}u0120\textbackslash{}""} & 369 & \texttt{"\textbackslash{}u0120for"} \\
354 & \texttt{"\textbackslash{}u2581for"} & 477 & \texttt{"\textbackslash{}u0120or"} & 594 & \texttt{"'s"} \\
369 & \texttt{"\textbackslash{}u2581that"} & 96618 & \texttt{"**:"} & 448 & \texttt{"\textbackslash{}u0120with"} \\
28731 & \texttt{")"} & 16 & \texttt{"1"} & 330 & \texttt{"\textbackslash{}u0120\textbackslash{}""} \\
395 & \texttt{"\textbackslash{}u2581with"} & 334 & \texttt{"**"} & 18 & \texttt{"3"} \\
9189 & \texttt{"**:"} & 649 & \texttt{"\textbackslash{}u0120can"} & 476 & \texttt{"\textbackslash{}u0120or"} \\
345 & \texttt{"\textbackslash{}u2581\textbackslash{}""} & 17 & \texttt{"2"} & 95518 & \texttt{"**:"} \\
28770 & \texttt{"3"} & 389 & \texttt{"\textbackslash{}u0120on"} & 334 & \texttt{"**"} \\
442 & \texttt{"\textbackslash{}u2581or"} & 439 & \texttt{"\textbackslash{}u0120as"} & 646 & \texttt{"\textbackslash{}u0120can"} \\
28730 & \texttt{"\_"} & 499 & \texttt{"\textbackslash{}u0120you"} & 389 & \texttt{"\textbackslash{}u0120on"} \\
28832 & \texttt{"`"} & 578 & \texttt{"\textbackslash{}u0120The"} & 438 & \texttt{"\textbackslash{}u0120as"} \\
541 & \texttt{"\textbackslash{}u2581can"} & 284 & \texttt{"\textbackslash{}u0120="} & 20 & \texttt{"5"} \\
356 & \texttt{"\textbackslash{}u2581on"} & 197 & \texttt{"\textbackslash{}u0109"} & 498 & \texttt{"\textbackslash{}u0120you"} \\
390 & \texttt{"\textbackslash{}u2581as"} & 701 & \texttt{"\textbackslash{}u0120your"} & 576 & \texttt{"\textbackslash{}u0120The"} \\
28782 & \texttt{"5"} & 18 & \texttt{"3"} & 19 & \texttt{"4"} \\
415 & \texttt{"\textbackslash{}u2581The"} & 262 & \texttt{"\textbackslash{}u0120\textbackslash{}u0120\textbackslash{}u0120"} & 284 & \texttt{"\textbackslash{}u0120="} \\
368 & \texttt{"\textbackslash{}u2581you"} & 1595 & \texttt{"\textbackslash{}u0120`"} & 197 & \texttt{"\textbackslash{}u0109"} \\
12 & \texttt{"<0x09>"} & 1473 & \texttt{":\textbackslash{}u010a\textbackslash{}u010a"} & 697 & \texttt{"\textbackslash{}u0120your"} \\
28781 & \texttt{"4"} & 527 & \texttt{"\textbackslash{}u0120are"} & 262 & \texttt{"\textbackslash{}u0120\textbackslash{}u0120\textbackslash{}u0120"} \\
327 & \texttt{"\textbackslash{}u2581="} & 433 & \texttt{"\textbackslash{}u0120it"} & 1565 & \texttt{"\textbackslash{}u0120`"} \\
1552 & \texttt{"\textbackslash{}u2581`"} & 8 & \texttt{")"} & 1447 & \texttt{":\textbackslash{}u010a\textbackslash{}u010a"} \\
574 & \texttt{"\textbackslash{}u2581your"} & 57277 & \texttt{"**\textbackslash{}u010a\textbackslash{}u010a"} & 525 & \texttt{"\textbackslash{}u0120are"} \\
288 & \texttt{"ing"} & 387 & \texttt{"\textbackslash{}u0120be"} & 432 & \texttt{"\textbackslash{}u0120it"} \\
227 & \texttt{"<0xE0>"} & 128001 & \texttt{"<|end\_of\_text|>"} & 8 & \texttt{")"} \\
28732 & \texttt{"("} & 1 & \texttt{"\textbackslash{}""} & 56177 & \texttt{"**\textbackslash{}u010a\textbackslash{}u010a"} \\
2287 & \texttt{"\textbackslash{}u2581\textbackslash{}u2581\textbackslash{}u2581"} & 63 & \texttt{"`"} & 387 & \texttt{"\textbackslash{}u0120be"} \\
28739 & \texttt{"\textbackslash{}""} & 358 & \texttt{"\textbackslash{}u0120I"} & 151645 & \texttt{"<|im\_end|>"} \\
460 & \texttt{"\textbackslash{}u2581are"} & 555 & \texttt{"\textbackslash{}u0120by"} & 1 & \texttt{"\textbackslash{}""} \\
378 & \texttt{"\textbackslash{}u2581it"} & 19 & \texttt{"4"} & 63 & \texttt{"`"} \\
347 & \texttt{"\textbackslash{}u2581be"} &  &  & 358 & \texttt{"\textbackslash{}u0120I"} \\
315 & \texttt{"\textbackslash{}u2581I"} &  &  & 553 & \texttt{"\textbackslash{}u0120by"} \\
2 & \texttt{"</s>"} &  &  &  &  \\
\end{longtable}
\endgroup

\subsection{Controlled Token Selection}
\label{app:adpo_token_selection_controls}

\begin{table}
\small
\setlength{\tabcolsep}{5pt}
\caption{Frequency selection and occurrence-matched randomized selection
on the training-calibration population. The range is the central empirical
range across randomized selections, not a confidence interval for a
trained-model effect. One-sided Monte Carlo values are unadjusted.}
\label{tab:app_adpo_occurrence_control}
\begin{tabular}{lrrrr}
\toprule
Statistic & Frequency & Random mean & Random 95\% range & $p$ \\
\midrule
Selected types & 52 & 20,911.4 & $[16{,}756.4,24{,}749.7]$ & --- \\
Occurrence coverage (\%) & 37.60 & 37.60 & $[37.60,37.60]$ & --- \\
\midrule
Shared coverage (\%) & 49.03 & 37.46 & $[34.62,40.71]$ & 0.001 \\
Shared enrichment & 1.304 & 0.996 & $[0.921,1.083]$ & 0.001 \\
Side specificity & 0.0152 & 0.0748 & $[0.0625,0.0864]$ & 0.001 \\
\bottomrule
\end{tabular}
\end{table}

\paragraph{An occurrence-budget control.}
Vocabulary cardinality and response-occurrence coverage describe different
properties of a selector. A fixed number of randomly selected vocabulary
types need not cover the same number of response positions as the most
frequent types. We therefore complement type-count comparisons with a
structural control that matches selected occurrence mass.

The Qwen2.5 control uses the first 4,000 training pairs of the same pinned
Llama-3-UltraFeedback release, the Qwen2.5-1.5B-Instruct tokenizer
(revision \texttt{989aa798}), chat formatting, response EOS, and the
2,048/1,800 sequence/prompt caps. At $\tau=0.002$, the frequency selector
contains 52 types and covers 1,194,145 of 3,176,166 response occurrences
(37.60\%). We draw 1,000 randomized selections from observed vocabulary
types without replacement, matching this occurrence budget rather than
the number of types. Specifically, each draw permutes the observed types
and accepts a type only when its complete occurrence count fits the
remaining budget; types that would exceed it are skipped. Selection stops
when the remaining budget reaches zero. Thus the controls are binary
type-level selections with exactly matched occurrence counts. Their
vocabulary sizes are allowed to vary. The randomization seed is 42.

We compare shared coverage $J$, occurrence-weighted side specificity
$\overline q$, and the shared enrichment
\begin{equation}
E(\mathcal S)=\frac{J(\mathcal S)}{F(\mathcal S)}.
\label{eq:app_adpo_shared_enrichment}
\end{equation}
An enrichment above one means that the selected occurrence mass contains
a disproportionately large share of chosen--rejected overlap.
The randomized selections define an empirical reference distribution for
each statistic. One-sided Monte Carlo values use the upper tail for
$J$ and $E$ and the lower tail for $\overline q$, with the plus-one
correction $(1+b)/(1+1{,}000)$, where $b$ counts randomized statistics
at least as extreme as the frequency selector in the tested direction.
At the matched occurrence budget, frequency selection captures more
shared chosen--rejected mass and substantially weaker token-identity side
specificity than randomized selection
(Table~\ref{tab:app_adpo_occurrence_control}). Its shared coverage is
49.03\%, compared with a random mean of 37.46\%, and its shared enrichment
is 1.304. Thus the observed structural concentration is not explained by
the amount of selected occurrence mass alone.

The held-out replication in
Appendix~\ref{app:adpo_frequency_structure} and this training-calibration
randomization address complementary questions: the former establishes
that the structural pattern reproduces on disjoint responses across
tokenizers, and the latter tests the specificity of frequency selection
at a fixed occurrence budget. Both analyses use token counts rather than
model optimization. They support frequency as a transparent selector of
shared, weakly side-specific response content for the score-masking
objective.